%% file: main.tex
\documentclass[letterpaper, 10 pt, conference]{ieeeconf}  % Comment this line out if you need a4paper

\IEEEoverridecommandlockouts                              % This command is only needed if 
\usepackage{graphicx}
\usepackage{amsmath,amssymb}
\usepackage{booktabs}
\usepackage{multirow}
\usepackage{array}
\usepackage[table]{xcolor}
\usepackage{microtype}
\usepackage{hyperref}
\definecolor{mydarkblue}{rgb}{0,0.08,0.45}
\hypersetup{
    pdfborder=0 0 0,
    colorlinks=true,
    linkcolor=mydarkblue,
    citecolor=mydarkblue,
    filecolor=mydarkblue,
    urlcolor=mydarkblue
}
\definecolor{darkgreen}{RGB}{0,155,0}
\title{\LARGE \bf
RMBench: Memory-Dependent Robotic Manipulation Benchmark with Insights into Policy Design}

\author{
Tianxing Chen$^{*1}$, Yuran Wang$^{*2,3}$, Mingleyang Li$^{*2}$, Yan Qin$^{*4}$, Hao Shi$^{5}$, Zixuan Li$^{6}$,\\
Yifan Hu$^{2}$, Yingsheng Zhang$^{5}$, Kaixuan Wang$^{1}$, Yue Chen$^{2}$, Hongcheng Wang$^{2}$,\\
Junjie Wang$^{5}$, Tianhang Yang$^{2}$, Renjing Xu$^{4}$, Ruihai Wu$^{2}$,\\
Yao Mu$^{7}$, Yaodong Yang$^{2,3}$, Hao Dong$^{\dagger 2}$, Ping Luo$^{\dagger 1}$\\
$*$ Equal contribution, $\dagger$ Corresponding authors\\
Website: \href{https://RMBench.github.io}{https://RMBench.github.io}\quad Code: \href{https://github.com/robotwin-Platform/rmbench}{https://github.com/robotwin-Platform/rmbench}%
\thanks{$^{1}$MMLab@HKU, $^{2}$PKU, $^{3}$PsiBot, $^{4}$HKUST (GZ), $^{5}$THU, $^{6}$SZU, $^{7}$SJTU.
Correspondence to: Tianxing Chen \textless chentianxing@connect.hku.hk\textgreater, Ping Luo \textless pluo@hku.hk\textgreater, Hao Dong \textless hao.dong@pku.edu.cn\textgreater.}
}

\begin{document}

\maketitle
\thispagestyle{empty}
\pagestyle{empty}

\input{sec/0-abstract}

\input{sec/1-intro}
\input{sec/2-related_work}
\input{sec/3-RMBench}
\input{sec/4-Mem-0}
\input{sec/5-experiment}
\input{sec/6-conclusion}
\bibliographystyle{IEEEtran}
\bibliography{IEEEfull,references}

\input{sec/7-appendix}

\end{document}

%% file: sec/0-abstract.tex
\begin{abstract}
Robotic manipulation policies have made rapid progress in recent years, yet most existing approaches give limited consideration to memory capabilities. Consequently, they struggle to solve tasks that require reasoning over historical observations and maintaining task-relevant information over time, which are common in real-world manipulation. Although several memory-aware policies have been proposed, systematic evaluation of memory-dependent manipulation remains underexplored, and the relationship between architectural choices and memory performance remains unclear. To address this gap, we introduce \textbf{RMBench}, a simulation benchmark of nine manipulation tasks spanning two levels of memory complexity for systematic evaluation of policy memory capabilities. We further propose \textbf{Mem-0}, a modular manipulation policy whose explicit memory components can be ablated independently. Evaluating eight policies on the nine simulated tasks and four policies on three physical tasks, we identify memory-related limitations in existing policies and show that the benefit of a memory component depends on whether the same information is already available elsewhere in the policy hierarchy, with subtask termination detection rather than memory representation forming the dominant bottleneck. Website: \href{https://rmbench-anonymous.github.io}{rmbench-anonymous.github.io}.\end{abstract}

%% file: sec/1-intro.tex
\section{Introduction}

Recent progress in robotic manipulation has demonstrated strong capabilities across a wide range of tasks. Modern policies such as Pi0.6~\cite{intelligence2025pi06vlalearnsexperience} and RDT2~\cite{rdt2} achieve impressive performance in flexible object manipulation and fine-grained skills, including complex activities like coffee making. Nevertheless, most existing robotic policies are primarily designed for short-horizon tasks and fine-grained manipulation. These approaches typically rely on a fixed-length window of recent observations, implicitly assuming that the underlying decision process is approximately Markovian.

In contrast, memory-dependent tasks are ubiquitous in real-world robotic applications. Such tasks are inherently non-Markovian, as past observations and actions may influence future decisions over extended temporal horizons. Examples include remembering the location of previously placed objects and reasoning over multiple attempts after forgetting a password. These tasks are both challenging and practically important, as they require robots to retain, retrieve, and utilize information beyond short-term sensory inputs.

Motivated by this challenge, several recent works have begun to explore memory-aware robotic policies. Approaches such as MemoryVLA~\cite{shi2025memoryvla}, MemER~\cite{sridhar2025memer}, CronusVLA~\cite{li2025cronusvla}, and SAM2Act~\cite{fang2025sam2act} incorporate explicit memory mechanisms to address long-horizon and memory-dependent decision making. Despite these efforts, the field currently lacks a systematically designed experimental platform for evaluating and analyzing robotic policies under long-term memory requirements. In particular, there is limited understanding of the underlying mechanisms that make memory strategies effective for robotic manipulation.

Existing benchmarks only partially bridge this gap. MemoryBench~\cite{fang2025sam2act} comprises seven single-arm manipulation tasks that involve memory, yet only three can be reliably reproduced in simulation, and the benchmark provides limited guidance on principled task design. MIKASA~\cite{cherepanov2025memory} introduces 32 memory-related manipulation tasks, but its formulation is largely tailored to reinforcement learning rather than general imitation learning. LIBERO-Long~\cite{liu2023liberobenchmarkingknowledgetransfer} offers ten long-horizon tasks, though these tasks do not explicitly demand memory since all task-relevant information remains observable throughout execution.

To address these limitations, we first introduce Task Memory Complexity, a principled metric for characterizing memory requirements in robotic manipulation tasks. This metric provides a systematic way to classify memory-dependent tasks and serves as a guideline for task design. Based on this formulation, we propose \textbf{RMBench}, a robotic manipulation benchmark built on the RoboTwin~2.0~\cite{chen2025robotwin} platform. RMBench consists of nine dual-arm manipulation tasks spanning different levels of task memory complexity, enabling large-scale and controlled studies of memory retention and utilization in robotic manipulation.

Furthermore, we propose \textbf{Mem-0}, a novel memory-oriented robotic policy designed with modular components that can be easily replaced or reconfigured. Mem-0 adopts a dual-system architecture in which a Subtask End Classifier detects the boundaries between task stages, allowing structured memory usage across long horizons. Through systematic ablation studies of Mem-0, we analyze which design components are critical for effective memory in robotic manipulation and derive insights for future policy design.

Our main contributions are summarized as follows:
\begin{itemize}
    \item We introduce Task Memory Complexity, a principled metric for categorizing memory-dependent robotic manipulation tasks, and propose \textbf{RMBench}, a simulation benchmark comprising nine memory-centric tasks based on this metric.
    \item We propose \textbf{Mem-0}, a memory-oriented robotic manipulation policy featuring a dual-system architecture with a Subtask End Classifier for flexible memory usage.
    \item We conduct comprehensive evaluations of representative state-of-the-art policies on RMBench and perform detailed ablation studies of Mem-0, revealing which policy design mechanisms are most beneficial for memory in robotic manipulation.
\end{itemize}

%% file: sec/2-related_work.tex
\section{Related Work}

\subsection{Robotic Manipulation Benchmarks}

Numerous simulation benchmarks underpin robotic manipulation research. RoboTwin~\cite{chen2025robotwin,mu2025robotwin}, RoboCasa~\cite{nasiriany2024robocasa}, ManiSkill3~\cite{tao2025maniskill3gpuparallelizedrobotics}, AutoBio~\cite{lan2025autobio}, UniVTAC~\cite{chen2026univtac}, DexGarmentLab~\cite{wang2025dexgarmentlab}, BEHAVIOR-1K~\cite{li2024behavior1k}, and SIMPLER~\cite{li2024evaluatingrealworldrobotmanipulation} offer diverse tasks, yet most are short-horizon and solvable without historical observations. Among memory-related benchmarks, MemoryBench~\cite{fang2025sam2act} contains few memory-dependent tasks and suffers from poor reproducibility and unclear design principles; MIKASA~\cite{cherepanov2025memory} offers a larger collection but is tailored to reinforcement learning; LIBERO-Long~\cite{liu2023liberobenchmarkingknowledgetransfer} and RoboCerebra~\cite{han2025robocerebra} are long-horizon yet keep task-relevant information observable throughout, so they do not explicitly require memory. In contrast, RMBench explicitly stratifies memory requirements, enabling systematic analysis of memory-based policies across different difficulty levels.

\subsection{Robotic Manipulation Policies}

Generative models and imitation learning have produced a wide range of manipulation policies with strong single-task performance~\cite{zhao2023learningfinegrainedbimanualmanipulation,chi2025diffusion,ze20243d,chen2025g3flow,lu2024manicm,su2025dense,chen2026mopacoordinatedmobilemanipulation}. Many adopt Vision-Language-Action (VLA) formulations~\cite{wen2025dexvla,lin2025hif,liang2025discrete,shen2025expertise,wen2025diffusionvla,wen2025tinyvla}, pretraining on large-scale robot data for improved generalization, with representatives including Pi0.5~\cite{intelligence2025pi05visionlanguageactionmodelopenworld}, Pi0.6~\cite{intelligence2025pi06vlalearnsexperience}, RDT2~\cite{rdt2}, and X-VLA~\cite{zheng2025xvlasoftpromptedtransformerscalable}; complementary efforts decouple high-level cognition from action synthesis~\cite{li2024cogact} or learn latent action world models that predict future dynamics~\cite{bi2025motus}. However, most existing policies rely on fixed-length observation histories and cannot selectively retain task-relevant information over long horizons. This motivates memory-aware approaches such as MemoryVLA~\cite{shi2025memoryvla}, MemER~\cite{sridhar2025memer}, CronusVLA~\cite{li2025cronusvla}, and SAM2Act~\cite{fang2025sam2act}, which aggregate, select, or consolidate past observations, while RoboMemory~\cite{lei2025robomemorybraininspiredmultimemoryagentic} organizes spatial, temporal, episodic, and semantic memory into a parallel multi-module architecture that conditions high-level planning rather than the low-level action decoder. Building on this line of work, we propose Mem-0, a modular memory-enabled policy for systematic ablation of memory components and architectural choices.

%% file: sec/3-RMBench.tex
\section{RMBench}

\begin{figure*}[tbp]
    \centering
    \includegraphics[width=1.0\textwidth]{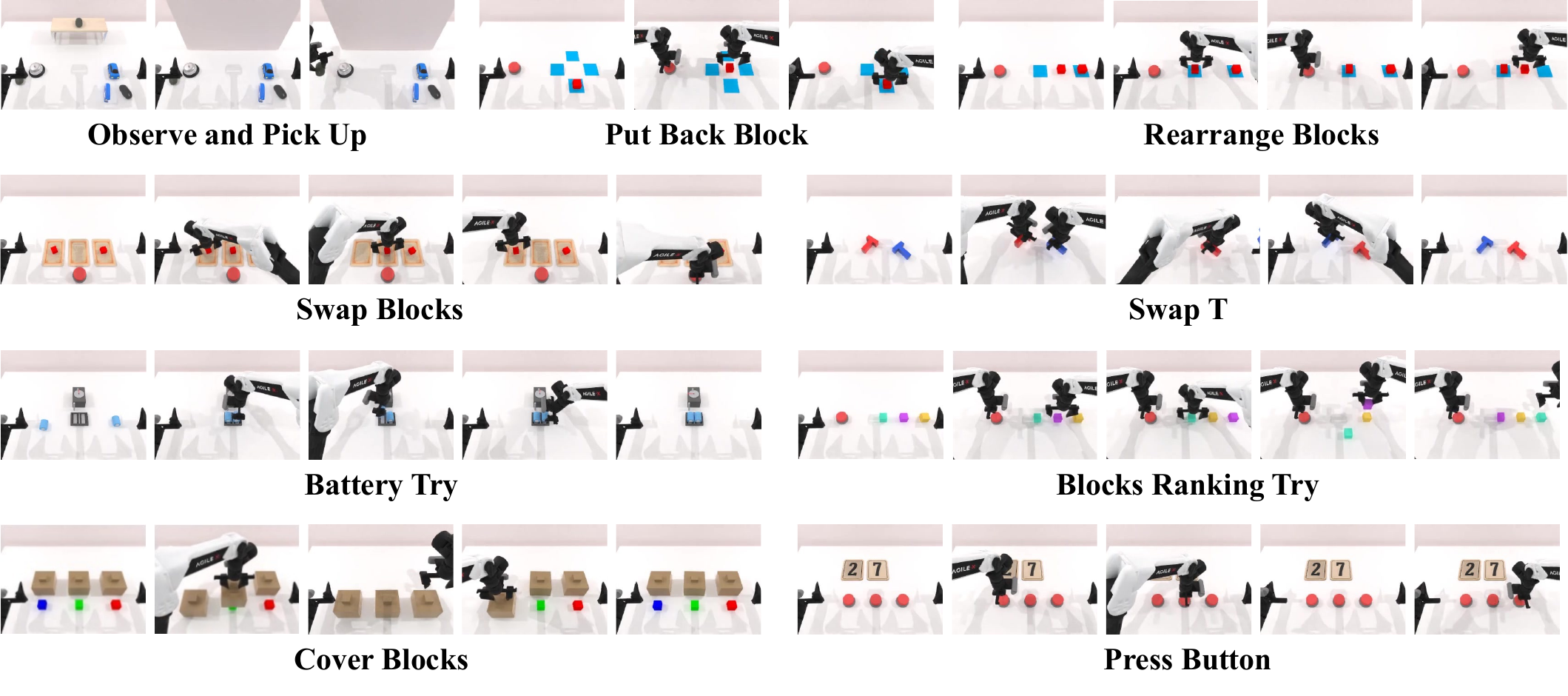}
    \vspace{-6mm}
        \caption{\textbf{RMBench Tasks.} We illustrate the nine memory-dependent tasks in RMBench together with their key execution steps. The $M(1)$ tasks include \emph{Observe and Pick Up}, \emph{Rearrange Blocks}, \emph{Put Back Block}, \emph{Swap Blocks}, and \emph{Swap T}; these tasks require retaining a single key observation to resolve state ambiguity across different execution stages. The $M(n)$ tasks include \emph{Battery Try}, \emph{Blocks Ranking Try}, \emph{Cover Blocks}, and \emph{Press Button}; these tasks involve repeated exploration, trial-and-error interaction, or task-specific event counting, and therefore require memory retention and retrieval over extended horizons.}
    \label{fig:rmbench_tasks}
    \vspace{-3mm}
\end{figure*}

In this section, we introduce \textbf{Task Memory Complexity}, a principled criterion for characterizing memory requirements in robotic manipulation tasks and guiding benchmark design. Based on this, we propose \textbf{RMBench}, a simulation benchmark comprising nine manipulation tasks with varying memory demands, designed to support controlled evaluation across different levels of task memory complexity.

\subsection{Task Memory Complexity (TMC)}
\label{sec:tmc}

Robotic manipulation frequently operates under partial observability: occlusions, delayed effects, and state aliasing make the current observation alone insufficient to determine task progress or the correct next action, and the information required to resolve this ambiguity generally lies not in a contiguous window of recent frames but in a few observations from arbitrary earlier time steps. Existing benchmarks assess memory only indirectly, through the measured performance of particular policy architectures, and therefore provide no task-centric criterion for how much past information a task actually requires. We address this gap with \textbf{Task Memory Complexity (TMC)}, which quantifies the minimum number of past observations that an optimal policy must retain.

\paragraph{Setup}
We model a manipulation task as a partially observable Markov decision process (POMDP) with latent state $s_t$, action $a_t$, and observation $o_t$. For the tasks considered here, the reward is the sparse success indicator specified by the task goal $g$, so that optimality (maximal expected discounted return) is well defined. The interaction history up to time $t$ is $h_t = (o_{1:t},\, a_{1:t-1})$.

\paragraph{Bounded memory}
Rather than characterizing memory through a particular learned representation, we characterize it through how many past observations a policy is permitted to access. A \emph{memory read of capacity} $k$ is a map
\begin{equation}
\mu_k(h_t) = \bigl( o_{\tau_1(h_t)},\, \ldots,\, o_{\tau_k(h_t)} \bigr),
\quad \tau_j(h_t) < t,
\label{eq:memread}
\end{equation}
where each selection rule $\tau_j$ may depend arbitrarily on $h_t$. A policy may thus choose \emph{which} $k$ past observations to retain, but not how many. Let $\mathcal{F}_k$ denote the set of all such maps, with $\mathcal{F}_0$ containing only the empty read. Because the selection rules are unconstrained, which observations matter is an outcome of the minimization below rather than an input to it.

\paragraph{Definition}
The \textbf{TMC} of a task is
\begin{equation}
\begin{split}
\mathrm{TMC} = \min \bigl\{\, & k \ge 0 \ \big| \ \exists\, \mu_k \in \mathcal{F}_k,\ \exists\, \pi \ \text{s.t.} \\[2pt]
& \pi\bigl( a_t \mid o_t, \mu_k(h_t), g \bigr) \ \text{is optimal} \,\bigr\}.
\end{split}
\label{eq:tmc}
\end{equation}
A task therefore satisfies $\mathrm{TMC} = k$ if and only if optimal behavior is attainable while retaining $k$ past observations but not fewer. Note that the policy in~\eqref{eq:tmc} conditions only on the current observation, the memory read, and the goal; it has access neither to the latent state $s_t$ nor to any privileged phase label.

\paragraph{Task annotation}
We annotate tasks as $M(0)$, $M(1)$, or $M(n)$ with $n \ge 2$, according to their TMC. $M(0)$ tasks are memory-free, in the sense that the current observation determines the optimal action; this is the regime targeted by most existing manipulation benchmarks, including the standard RoboTwin~2.0 task suite. $M(1)$ tasks require a single past observation to disambiguate the current state. $M(n)$ tasks require two or more non-local observations or event records, and consequently exhibit multi-step temporal dependencies. RMBench deliberately populates the two memory-dependent levels, $M(1)$ and $M(n)$.

\paragraph{Practical estimation of TMC}
Although observations are continuous in time, TMC is a task-level quantity rather than a discretization of physical time. The selection rules $\tau_j$ index semantic events, such as an initially observed configuration, a previous placement attempt, a failed insertion trial, or the number of completed button presses, so $n$ counts distinct semantic memories rather than raw frames. In practice we estimate TMC by enumerating the decision points at which the current observation admits more than one admissible action, identifying for each such point the past observations that resolve the ambiguity, and then discarding redundant items. The number of remaining independent items is the TMC of the task.

\paragraph{Invariance to task decomposition}
Any long-horizon task can be made to appear memory-free by decomposing it into primitives and supplying oracle phase labels. Definition~\eqref{eq:tmc} excludes this degenerate case by construction, because the policy is not permitted to condition on phase labels. Equivalently, TMC is annotated at the coarsest level at which the task goal $g$ is expressed as a single instruction. For example, \textit{Put Back Block} requires the initial pad of the block to be recalled after the block has been moved to the center and a button has been pressed: the trajectory can be split into visually guided primitives, yet the terminal placement still depends on one earlier observation, so the task is annotated as $M(1)$. \textit{Blocks Ranking Try}, in contrast, requires the outcomes of several previous attempts to be accumulated, and is therefore annotated as $M(n)$.

\subsection{RMBench System Design}
\label{sec:system_design}

Based on TMC, we design nine memory-dependent manipulation tasks, including five $M(1)$ tasks and four $M(n)$ tasks, with representative key frames shown in Fig.~\ref{fig:rmbench_tasks}. RMBench is developed within the RoboTwin~2.0~\cite{chen2025robotwin} framework and built on the SAPIEN~\cite{xiang2020sapien} simulation engine, which supports automated data synthesis and integrated policy evaluation in a unified pipeline. This design enables scalable demonstration generation as well as consistent and reproducible benchmarking of manipulation policies. In addition, RMBench provides fine-grained language annotations aligned with each action--observation pair. These annotations assign explicit linguistic descriptions to low-level interactions and state transitions, and supply the subtask-level supervision used by the hierarchical policies evaluated in Sec.~\ref{sec:exp_on_baseline}. Because this supervision is a property of the benchmark rather than of any single method, it is available to all policies under evaluation.

% Sourced here (one page ahead) so that the double-column float lands on the
% same page as Sec. IV (Mem-0 Policy); it belongs to that section.
\begin{figure*}[t]
    \centering
    \includegraphics[width=1.0\textwidth]{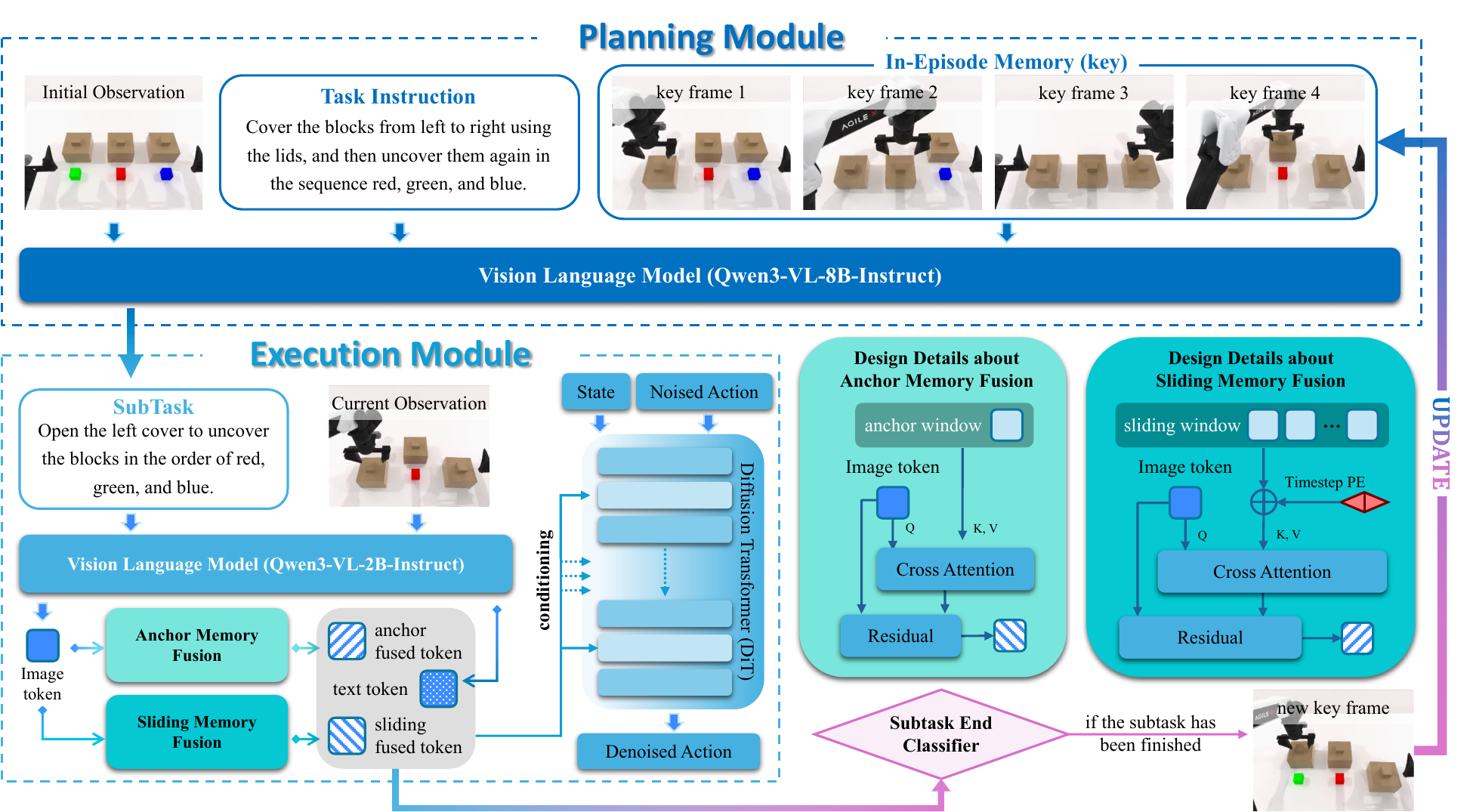}
    \caption{\textbf{Mem-0 Pipeline.} Mem-0 comprises a Planning Module and an Execution Module linked by a Subtask End Classifier. The Planning Module generates high-level subtasks from the task instruction, the initial observation, and the key memory window (In-Episode Memory), while the Execution Module produces low-level actions from the current observation, subtask, and fused anchor and sliding memories through a flow-matching policy. Upon subtask completion, a key frame is appended to the key memory window, enabling iterative planning and execution until the task terminates.}
    \label{fig:mem0}
    \vspace{-3mm}
\end{figure*}

%% file: sec/4-Mem-0.tex
\section{Mem-0 Policy}

In this section, we present \textbf{Mem-0}, a modular memory-oriented robotic policy designed for systematic analysis of memory mechanisms. As shown in Fig.~\ref{fig:mem0}, Mem-0 consists of a \textbf{Planning Module} (Sec.~\ref{sec:planning}) that performs subtask-level reasoning over a key memory window, and an \textbf{Execution Module} (Sec.~\ref{sec:execution}) that executes the current subtask using an anchor memory and a sliding memory window. The two modules are connected by a \textbf{Subtask End Classifier} (Sec.~\ref{sec:classifier}), which enables closed-loop planning and execution.

\subsection{Planning Module}
\label{sec:planning}

In long-horizon manipulation, end-to-end VLA models are prone to error accumulation and trajectory drift, which prior work~\cite{sridhar2025memer,wen2025diffusionvla} mitigates through subtask decomposition. Such approaches remain insufficient in memory-dependent settings such as the $M(n)$-type tasks in RMBench, where inferring the next subtask requires reasoning over multiple previously completed ones, for example by tracking the number of button presses. We therefore introduce a \emph{key memory window} $\mathcal{M}_{\mathrm{K}}$ that aggregates completed subtasks and enables memory-aware subtask reasoning.

Concretely, a vision-language model $\mathcal{V}_{\text{plan}}$ predicts the $i$-th subtask as $s_i = \mathcal{V}_{\text{plan}}\!\left(o_0,\, g,\, \mathcal{M}_{\mathrm{K}}^{i-1}\right)$, where $o_0$ is the episode-initial RGB observation, $g$ is the natural-language task goal, and $\mathcal{M}_{\mathrm{K}}^{i-1} = \{( s_j,\, o_j^{\mathrm{end}} )\}_{j=1}^{i-1}$ collects the textual description $s_j$ and the terminal observation $o_j^{\mathrm{end}}$ of every completed subtask. Conditioning on $\mathcal{M}_{\mathrm{K}}^{i-1}$ lets the model reason explicitly over previously executed subtasks and their visual outcomes, beyond single-frame planning.

Unlike approaches that infer a new subtask at every frame and thus require $O(T)$ planning calls over a $T$-timestep horizon, the Planning Module is invoked only upon subtask termination, as detected by the Subtask End Classifier (Sec.~\ref{sec:classifier}); for a task with $N \ll T$ subtasks this requires only $O(N)$ invocations, so the Execution Module runs at its native control rate without being throttled by planning latency.

\subsection{Execution Module}
\label{sec:execution}

Although subtask-level planning is effective for many memory-dependent tasks, some tasks do not admit a useful subtask decomposition: in $M(1)$-type tasks such as \emph{Swap T}, the target placement orientation of the T-shaped object cannot be specified reliably in language, and overly fine-grained decomposition increases annotation effort and planning latency. To address these limitations, we incorporate an \textbf{anchor memory} $\mathcal{M}_{\mathrm{A}}$ and a \textbf{sliding memory window} $\mathcal{M}_{\mathrm{S}}$ within the Execution Module, which together allow $M(1)$-type tasks to be handled without any subtask decomposition by maintaining one persistent memory slot alongside a short-term buffer.

The Execution Module executes the current subtask with a flow-matching policy conditioned on multimodal perception and memory. A vision-language model $\mathcal{V}_{\text{exec}}$ encodes the current observation $o_t$ and the active subtask instruction $s_i$ into image and text token embeddings, and mean pooling yields the compact latents $\mathbf{z}_t^{\text{img}}$ and $\mathbf{z}_t^{\text{text}}$. The image latent is then read against the two memory buffers with a shared residual cross-attention operator, using $\mathbf{z}_t^{\text{img}}$ as the query and the buffer contents as keys and values,
\begin{equation}
    \tilde{\mathbf{z}}_t^{\,l}
    =
    \mathrm{CrossAttn}\!\left(
    \mathbf{z}_t^{\text{img}},\;
    \mathcal{B}^{\,l}_t
    \right) + \mathbf{z}_t^{\text{img}},
    \label{eq:crossattn}
\end{equation}
for $l \in \{\text{anchor}, \text{slide}\}$, where $\mathcal{B}^{\text{anchor}}_t = \mathcal{M}_{\mathrm{A}}$ and $\mathcal{B}^{\text{slide}}_t = \mathcal{M}_{\mathrm{S}}^{t}$. For the sliding read, a learned timestep positional encoding is added to the keys and values so that the policy can distinguish recent from older entries; the anchor read needs none, since its content is by construction the earliest latent of the subtask. The two fused tokens are concatenated with the text token into the conditioning vector $\mathbf{c}_t = [\,\tilde{\mathbf{z}}_t^{\text{anchor}};\, \tilde{\mathbf{z}}_t^{\text{slide}};\, \mathbf{z}_t^{\text{text}}\,]$.

The two buffers are maintained differently. The sliding window keeps the $K$ most recent image latents, $\mathcal{M}_{\mathrm{S}}^{t+1} = \mathrm{Trunc}_K\bigl(( \mathcal{M}_{\mathrm{S}}^{t},\, \mathbf{z}_t^{\text{img}} )\bigr)$, where $\mathrm{Trunc}_K(\cdot)$ discards all but the $K$ most recent elements. The anchor memory holds a single slot, $\mathcal{M}_{\mathrm{A}} = \{\mathbf{z}_{t_i}^{\text{img}}\}$, fixed at the image latent of the first control timestep $t_i$ of subtask $i$ and left unchanged while that subtask executes. When a subtask transition is declared (Sec.~\ref{sec:classifier}), both buffers are cleared, $\mathcal{M}_{\mathrm{A}} \leftarrow \varnothing$ and $\mathcal{M}_{\mathrm{S}} \leftarrow \varnothing$, so that memory persisting across subtasks is carried exclusively by the key memory window.

The two reads are not symmetric, which bears directly on the ablations in Sec.~\ref{sec:exp_on_ablation}. Because $|\mathcal{M}_{\mathrm{A}}| = 1$, the attention weights of the anchor read are identically one, and~\eqref{eq:crossattn} reduces to a learned projection of the subtask-initial latent added to the current latent: the anchor read performs no selection, and its benefit stems from the persistence of that context rather than from retrieval. The sliding read, in contrast, attends over $K$ entries and performs genuine temporal selection. Enlarging the anchor buffer to support selective retrieval is left to future work.

For action generation we employ a diffusion transformer (DiT) trained with a flow-matching objective, with action horizon $H = 30$ over the $d_a = 16$-dimensional bimanual action space. Let $\mathbf{a} \in \mathbb{R}^{H \times d_a}$ denote the ground-truth action chunk $\mathbf{a}_{t:t+H-1}$. During training we draw $\epsilon \sim \mathcal{N}(\mathbf{0}, \mathbf{I})$ and a flow time $\tau$, interpolate linearly between noise and data, and predict the velocity of the path from the proprioceptive state $\mathbf{x}_t$ and the memory-conditioned vector $\mathbf{c}_t$,
\begin{equation}
    \mathbf{a}^{\tau} = (1-\tau)\,\epsilon + \tau\,\mathbf{a},
    \quad
    \hat{\mathbf{v}} = \mathrm{DiT}\!\left(\mathbf{a}^{\tau},\, \tau,\, \mathbf{x}_t,\, \mathbf{c}_t\right),
    \label{eq:flow}
\end{equation}
with regression target $\mathbf{v} = \mathbf{a} - \epsilon$. Flow times follow a shifted Beta distribution, $\tau = (s - u)/s$ with $u \sim \mathrm{Beta}(1.5, 1.0)$ and $s = 0.999$, which concentrates supervision near the data end of the path. At deployment the chunk is produced by Euler integration of~\eqref{eq:flow} from $\epsilon$ over eight steps, and multiple actions from each chunk are executed before the next inference call, sustaining 5--10~Hz control on a single NVIDIA A800 GPU.

\subsection{Subtask End Classifier}
\label{sec:classifier}

To close the loop between the Planning and Execution Modules, we introduce a \emph{Subtask End Classifier} that detects subtask completion. A lightweight multilayer perceptron maps the conditioning vector $\mathbf{c}_t$ to a logit $\ell_t = \mathrm{MLP}(\mathbf{c}_t)$, thresholded into the binary termination signal $\mathcal{C}_{\text{end}}(\mathbf{c}_t) = \mathbf{1}\!\left[\sigma(\ell_t) \ge 0.5\right]$, where $\sigma$ is the logistic function; because it shares $\mathbf{c}_t$ with the action head, the classifier consumes no privileged simulator information. Acting on a single positive prediction is not robust, so we accumulate the count $n_t = \sum_{j = t_i}^{t} \mathcal{C}_{\text{end}}(\mathbf{c}_j)$ of termination signals raised since the current subtask began, and declare a transition once $n_t \ge L_{\text{sw}}$ with $L_{\text{sw}} = 8$, which suppresses isolated false positives while keeping transitions responsive. On a transition the counter is reset and both buffers are cleared (Sec.~\ref{sec:execution}); the same reset is applied during training, except that the $L_{\text{mem}} = 8$ termination frames that trigger it are read from the ground-truth annotations rather than predicted. Upon transition, the final observation $o_i^{\mathrm{end}} = o_t$ and the subtask description $s_i$ are appended to $\mathcal{M}_{\mathrm{K}}^{i}$ and passed to the Planning Module for the next round of subtask-level reasoning.

\subsection{Training Objective}
\label{sec:objective}

The Execution Module and the Subtask End Classifier are trained jointly on expert demonstrations $\mathcal{D}$, with $y_t \in \{0,1\}$ the ground-truth termination label derived from the language annotations of Sec.~\ref{sec:system_design}:
\begin{equation}
    \mathcal{L}
    =
    \underbrace{\mathbb{E}\!\left[\| \hat{\mathbf{v}} - \mathbf{v} \|_2^2\right]}_{\mathcal{L}_{\text{act}}}
    +
    \lambda\,
    \underbrace{\mathbb{E}\!\left[(1 - p_t)^{\gamma_{\mathrm{f}}}\, \mathrm{BCE}_{w}( \ell_t,\, y_t )\right]}_{\mathcal{L}_{\text{end}}}.
    \label{eq:loss}
\end{equation}
Here $\mathcal{L}_{\text{act}}$ is the flow-matching regression of~\eqref{eq:flow} onto the target velocity, with expectations taken over $\mathcal{D}$ and the sampled $(\epsilon, \tau)$. In $\mathcal{L}_{\text{end}}$, $\mathrm{BCE}_{w}$ is binary cross-entropy on the logit $\ell_t$ with a positive-class weight $w$ that compensates for the rarity of termination frames, $p_t$ is the predicted probability of the ground-truth class, and the focal exponent $\gamma_{\mathrm{f}}$ down-weights easy negatives; the scalar $\lambda$ balances the two terms. The Planning Module is fine-tuned separately with a standard autoregressive language-modeling loss on subtask descriptions.

%% file: sec/5-experiment.tex
\input{tables/benchmark}

\section{Experiment}

We design a set of experiments to validate three key objectives:
(1) to evaluate the performance of existing manipulation policies and Mem-0 on RMBench, thereby characterizing their ability to handle memory-dependent tasks across different levels of difficulty;
(2) to conduct systematic ablation studies on the Mem-0 architecture in order to analyze how different module designs affect performance on memory-intensive manipulation tasks; and
(3) to perform real-world robotic experiments to assess the effectiveness and generalization of Mem-0 beyond simulation.

\subsection{Evaluation of Policies on RMBench}
\label{sec:exp_on_baseline}

We benchmark eight policies on RMBench in three groups: DP and ACT trained from scratch, the large-scale pretrained Pi0.5, X-VLA, and QwenOFT, and the memory-based MemER, MemoryVLA, and our Mem-0. QwenOFT and MemoryVLA are implemented with StarVLA~\cite{community2026starvlalegolikecodebasevisionlanguageaction} (MemoryVLA on the OpenVLA backbone), and all policies are trained and evaluated through the unified interface of XPolicyLab~\cite{community2026xpolicylab}, which our released code follows. Since MemER was developed for a single arm, we adapt its action space to the dual-arm setting of RoboTwin~2.0 while leaving its memory mechanism unchanged. Each method is trained on the same 50 expert demonstrations per task (with the objective of Sec.~\ref{sec:objective} for Mem-0) and evaluated over 100 rollout episodes; Table~\ref{tab:benchmark} reports success rates.

Two protocol aspects shape how the comparison should be read. First, Mem-0 and MemER are hierarchical and consume the subtask annotations and termination labels provided by RMBench, whereas the remaining policies are trained end-to-end from observation--action pairs; this supervision is a property of the benchmark and available to any policy, but the gap between Mem-0 and the end-to-end baselines reflects both the memory architecture and this additional training signal. Second, Mem-0 runs without subtask decomposition on $M(1)$ tasks, isolating the Execution Module, and with decomposition on $M(n)$ tasks, reflecting the joint behavior of both modules; Sec.~\ref{sec:exp_on_ablation} separates these contributions.

\begin{figure*}[t]
    \centering
    \vspace{1mm}
    \includegraphics[width=1.0\textwidth]{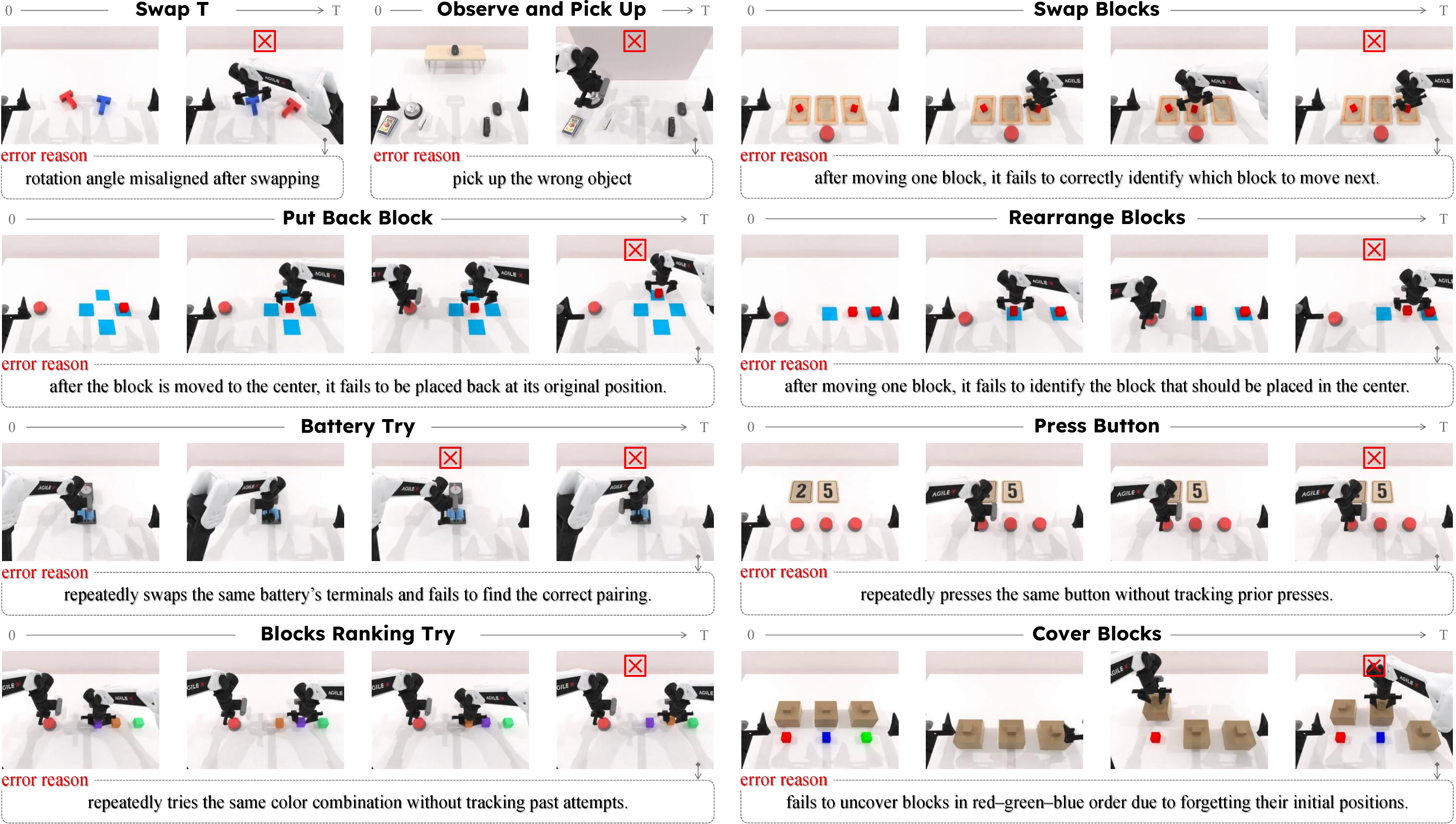}
    \vspace{-5mm}
    \caption{\textbf{Error Analysis.} Representative failure mode and its cause on each of the nine RMBench tasks.}
    \label{fig:error_analysis}
    \vspace{-5mm}
\end{figure*}

Policies without explicit memory perform poorly across RMBench, confirming their difficulty on non-Markovian tasks where the current observation alone is insufficient; Fig.~\ref{fig:error_analysis} summarizes the dominant failure mode on each task (see also the supplementary video). Averaged over tasks, Mem-0 improves success rate by 33.2 percentage points on $M(1)$ tasks and 9.2 points on $M(n)$ tasks over the strongest baseline, MemoryVLA. The asymmetry is informative: $M(1)$ tasks need only a single episode-initial observation, which the anchor memory captures directly, whereas $M(n)$ tasks depend on an accumulated event history routed through subtask-level reasoning, leaving more room for error.

Mem-0 nevertheless remains far from solving RMBench. On \textit{Observe and Pick Up} all methods stay below 10\% and pretrained policies keep a small edge: the bottleneck is fine-grained semantic discrimination, not memory. On \textit{Swap T}, performance is limited by the precision needed to match the target orientation. On \textit{Press Button} every policy scores zero, since a single press produces only a subtle visual change that defeats termination detection.

\subsection{Analysis of Memory-Related Modules}
\label{sec:exp_on_ablation}

\input{tables/ablation}

We ablate Mem-0 in four ways: \textbf{w/o Anchor} and \textbf{w/o Sliding} remove the corresponding Execution-Module buffer; \textbf{w/o Key} removes the key memory window, so subtask inference conditions only on $o_0$ and $g$; and \textbf{GT Classifier} replaces the learned Subtask End Classifier with ground-truth termination signals from the simulator, reported as an oracle rather than a competing configuration since it consumes privileged information unavailable at deployment. Because Mem-0 uses no subtask decomposition on $M(1)$ tasks, only the first two apply there. Tables~\ref{tab:ablation} and~\ref{tab:ablation_mn} report the $M(1)$ and $M(n)$ results.

\textbf{Anchor memory.}
On $M(1)$ tasks the anchor is decisive: removing it lowers the average from 52.8\% to 26.8\%, most sharply on \textit{Put Back Block} (90\% to 35\%) and \textit{Swap Blocks} (67\% to 15\%). This is expected, since an $M(1)$ episode runs as a single subtask whose anchor slot holds exactly the episode-initial configuration the task requires. On $M(n)$ tasks the picture partly reverses: the average barely moves (28.5\% to 26.8\%), and on \textit{Cover Blocks} removal in fact \emph{improves} success from 68\% to 92\%. Under subtask decomposition the needed memory is carried by the key window, as confirmed by \textit{Cover Blocks} collapsing to 5\% under \textbf{w/o Key}; the per-subtask anchor thus adds no information and instead biases the policy toward the scene at subtask onset, which is actively harmful when a subtask requires closed-loop alignment with a lid whose pose keeps changing. The conclusion is narrower than a blanket endorsement of persistent memory: anchor memory helps when it is the sole carrier of task-relevant history, and hurts when it duplicates higher-level memory.

\textbf{Sliding memory.}
Removing the sliding window reduces the $M(1)$ average from 52.8\% to 40.4\% and the $M(n)$ average from 28.5\% to 25.3\%, again not uniformly: the ablated model is better on \textit{Swap T} (20\% versus 14\%), where sensitivity to the initial orientation and fairly stereotyped motion favor shifting reliance onto the anchor slot, and on \textit{Cover Blocks} (84\% versus 68\%), consistent with the interference above. Together, the two ablations show that the marginal value of each buffer depends on whether the same information is already available elsewhere in the hierarchy, and that naively adding memory pathways can degrade performance.

\textbf{Key memory.}
Removing the key window is unambiguously damaging: the $M(n)$ average falls from 28.5\% to 4.8\%, with \textit{Blocks Ranking Try} dropping from 18\% to 1\% and \textit{Cover Blocks} from 68\% to 5\%. Without completed subtasks and their visual outcomes, the Planning Module must infer the next subtask from the episode-initial observation alone, which cannot disambiguate how many attempts have been made or which blocks have been handled. Key memory thus supplies the multi-step history that $M(n)$ tasks actually require, and is the only component whose removal degrades every $M(n)$ task on which any method makes progress.

\textbf{Subtask End Classifier.}
The classifier keeps planning at $O(N)$ calls, but its accuracy is the dominant bottleneck of the current system: with ground-truth termination signals the $M(n)$ average rises from 28.5\% to 45.3\% (\textit{Blocks Ranking Try} 18\% to 45\%, \textit{Press Button} 0\% to 14\%). Since the oracle changes nothing about the memory representation, this 16.8-point gap stems entirely from mistimed subtask transitions, and it exceeds the gap between Mem-0 and the strongest baseline on the same tasks. We therefore regard robust subtask termination detection, rather than richer memory representation, as the most consequential direction for improving hierarchical memory-dependent policies.

\subsection{Real-World Experiments}

\begin{figure}[htbp]
    \centering
    \includegraphics[width=\columnwidth]{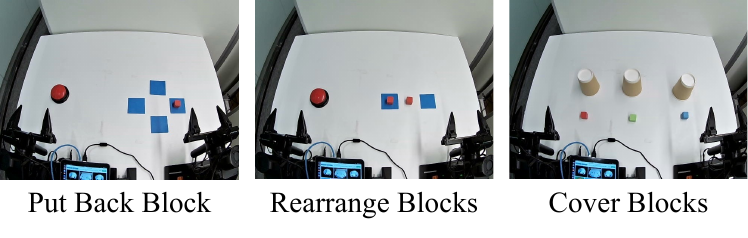}
    \vspace{-8mm}
    \caption{\textbf{Real-world Tasks Visualization.} Setups for the three physical tasks used in the real-world evaluation.}
    \label{fig:real-world}
    \vspace{-1mm}
\end{figure}

We evaluate Mem-0 on three physical tasks aligned with RMBench, \textit{Put Back Block}, \textit{Rearrange Blocks}, and \textit{Cover Blocks}, against DP, ACT, and Pi0.5, with all data collected on the X-One dual-arm platform (Fig.~\ref{fig:real-world}). Each policy is trained on 100 teleoperated demonstrations per task and evaluated over 40 rollout trials; Table~\ref{tab:real-world} reports success rates.

% ---------------- Figure ----------------

\input{tables/real_world}

Mem-0 attains the highest success rate on all three tasks, averaging 22.5\% against 5.8\% for Pi0.5, while DP and ACT complete no trial. Given the low absolute values, we read this experiment as evidence that the memory mechanism transfers to hardware rather than as a demonstration of deployable reliability. Most Mem-0 failures stem from imprecise block manipulation rather than incorrect high-level planning, consistent with the simulation finding that memory is not the limiting factor for low-level accuracy; we attribute the low absolute performance to the higher behavioral variability of teleoperated demonstrations and to training from scratch without manipulation pretraining. Two limitations remain: the tasks cover only the $M(1)$ and $M(n)$ patterns that are straightforward to instrument physically, and no memory-based baseline is included, as neither MemER nor MemoryVLA was deployable on our platform.

%% file: tables/benchmark.tex
\begin{table*}[t]
\centering
\renewcommand{\arraystretch}{0.85}
\setlength{\tabcolsep}{2pt}
\scriptsize
\caption{\textbf{RMBench benchmark results.}
Success rates on the nine RMBench tasks spanning $M(1)$ and $M(n)$.
Total Average is the unweighted mean over all nine tasks.
(\colorbox{blue!15}{\textbf{Bold}}: best; \underline{Underlined}: second-best; \textcolor{darkgreen}{Green}: gain over the second-best).
}
\resizebox{0.95\textwidth}{!}{%
\begin{tabular}{lc|ccccc|cc|c}
\toprule
Tasks & TMC & DP & ACT & Pi0.5 & X-VLA & QwenOFT & MemER & MemoryVLA & Mem-0 (ours) \\
\midrule
Observe and Pick Up & $M(1)$ & 1\% & 1\% & \cellcolor{blue!15}\textbf{9\%} & \cellcolor{blue!15}\textbf{9\%} & 0\% & \underline{7\%} & 0\% & 4\% \\
Rearrange Blocks & $M(1)$ & 0\% & \underline{29\%} & 13\% & 13\% & 0\% & 17\% & 22\% & \cellcolor{blue!15}\textbf{89\%} \\
Put Back Block & $M(1)$ & 0\% & 0\% & 11\% & 18\% & 0\% & 0\% & \underline{50\%} & \cellcolor{blue!15}\textbf{90\%} \\
Swap Blocks & $M(1)$ & 11\% & 2\% & \underline{24\%} & 16\% & 0\% & 14\% & 17\% & \cellcolor{blue!15}\textbf{67\%} \\
Swap T & $M(1)$ & \cellcolor{blue!15}\textbf{20\%} & 2\% & \underline{15\%} & 3\% & 0\% & 7\% & 9\% & 14\% \\
\textbf{\textit{Average}} & $M(1)$ & 6.4\% & 6.8\% & 14.4\% & 11.8\% & 0.0\% & 9.0\% & \underline{19.6\%} & \cellcolor{blue!15}\textbf{52.8\%\,(\textcolor{darkgreen}{\footnotesize +33.2\%})} \\
\midrule
Battery Try & $M(n)$ & 10\% & 19\% & 16\% & 26\% & 14\% & \underline{27\%} & 25\% & \cellcolor{blue!15}\textbf{28\%} \\
Blocks Ranking Try & $M(n)$ & 10\% & 0\% & 6\% & 1\% & \cellcolor{blue!15}\textbf{37\%} & 0\% & 12\% & \underline{18\%} \\
Cover Blocks & $M(n)$ & 0\% & 0\% & 0\% & 2\% & 0\% & 6\% & \underline{40\%} & \cellcolor{blue!15}\textbf{68\%} \\
Press Button & $M(n)$ & 0\% & 0\% & 0\% & 0\% & 0\% & 0\% & 0\% & 0\% \\
\textbf{\textit{Average}} & $M(n)$ & 5.0\% & 4.8\% & 5.5\% & 7.3\% & 12.8\% & 8.3\% & \underline{19.3\%} & \cellcolor{blue!15}\textbf{28.5\%\,(\textcolor{darkgreen}{\footnotesize +9.2\%})} \\
\midrule
\textit{\textbf{Total Average}} & / & 5.8\% & 5.9\% & 10.4\% & 9.8\% & 5.7\% & 8.7\% & \underline{19.4\%} & \cellcolor{blue!15}\textbf{42.0\%\,(\textcolor{darkgreen}{\footnotesize +22.6\%})} \\
\bottomrule
\end{tabular}%
}
\label{tab:benchmark}
\end{table*}

%% file: tables/ablation.tex
\begin{table}[t]
\centering
\renewcommand{\arraystretch}{0.95}
\setlength{\tabcolsep}{3pt}
\small
\caption{\textbf{Ablation Studies on $M(1)$ Tasks.} (\colorbox{blue!15}{\textbf{Bold}}: the best results; \underline{Underlined}: the second-best results).}
\label{tab:ablation}
\resizebox{0.9\columnwidth}{!}{%
\begin{tabular}{l|ccc}
\toprule
$M(1)$ Tasks & Mem-0 (ours) & w/o Anchor & w/o Sliding \\
\midrule
Observe and Pick Up & \cellcolor{blue!15}\textbf{4\%} & \cellcolor{blue!15}\textbf{4\%} & \underline{3\%} \\
Rearrange Blocks & \cellcolor{blue!15}\textbf{89\%} & \underline{73\%} & 62\% \\
Put Back Block & \cellcolor{blue!15}\textbf{90\%} & 35\% & \underline{78\%} \\
Swap Blocks & \cellcolor{blue!15}\textbf{67\%} & 15\% & \underline{39\%} \\
Swap T & \underline{14\%} & 7\% & \cellcolor{blue!15}\textbf{20\%} \\
\midrule
\textbf{\textit{Average}} & \cellcolor{blue!15}\textbf{52.8\%} & 26.8\% & \underline{40.4\%} \\
\bottomrule
\end{tabular}%
}
\end{table}

\vspace{6pt}

\begin{table}[t]
\centering
\renewcommand{\arraystretch}{0.95}
\setlength{\tabcolsep}{3pt}
\small
\caption{\textbf{Ablation Studies on $M(n)$ Tasks.} GT Classifier is an oracle and excluded from the ranking (\colorbox{blue!15}{\textbf{Bold}}: best deployable configuration; \underline{Underlined}: second-best).}
\label{tab:ablation_mn}
\resizebox{\columnwidth}{!}{%
\begin{tabular}{l|cccc|c}
\toprule
$M(n)$ Tasks & Mem-0 (ours) & w/o Key & w/o Anchor & w/o Sliding & \textit{GT Classifier} \\
 & & & & & \textit{(oracle)} \\
\midrule
Battery Try & \cellcolor{blue!15}\textbf{28\%} & 13\% & 14\% & \underline{17\%} & \textit{30\%} \\
Blocks Ranking Try & \cellcolor{blue!15}\textbf{18\%} & \underline{1\%} & 0\% & 0\% & \textit{45\%} \\
Cover Blocks & 68\% & 5\% & \cellcolor{blue!15}\textbf{92\%} & \underline{84\%} & \textit{92\%} \\
Press Button & 0\% & 0\% & \cellcolor{blue!15}\textbf{1\%} & 0\% & \textit{14\%} \\
\midrule
\textbf{\textit{Average}} & \cellcolor{blue!15}\textbf{28.5\%} & 4.8\% & \underline{26.8\%} & 25.3\% & \textit{45.3\%} \\
\bottomrule
\end{tabular}%
}
\vspace{-2mm}
\end{table}

%% file: tables/real_world.tex
% ---------------- Table ----------------
\begin{table}[htbp]
    \centering
    \small
    \setlength{\tabcolsep}{5pt}
    \vspace{-3mm}
    \caption{\textbf{Real-world Evaluation Results.} Success rates on the three physical tasks (\textbf{Bold}: best).}

    \label{tab:real-world}

    \begin{tabular}{lcccc}
        \toprule
        Tasks & DP & ACT & Pi0.5 & Mem-0 (ours) \\
        \midrule

        Put Back Block   & 0.0\%  & 0.0\% & 10.0\% & \textbf{17.5\%} \\
        Rearrange Blocks & 0.0\%  & 0.0\% & 7.5\%  & \textbf{37.5\%} \\
        Cover Blocks     & 0.0\%  & 0.0\% & 0.0\%  & \textbf{12.5\%} \\

        \midrule
        \textit{\textbf{Average}} & 0.0\% & 0.0\% & 5.8\% & \textbf{22.5\%} \\

        \bottomrule
    \end{tabular}
\end{table}

%% file: sec/6-conclusion.tex
\section{Conclusion}

In this paper, we present RMBench (benchmark) and Mem-0 (policy) to systematically evaluate memory in robotic manipulation, revealing the memory limitations of existing policies and how different architectural choices (such as anchor memory, sliding memory, and key memory) affect memory performance, thereby providing preliminary insights into the principled integration of memory mechanisms for effective memory-dependent robotic manipulation. As for future work, promising directions include improved memory representation and fusion, more robust subtask termination criteria, and pretraining to enhance semantic understanding and generalization. We hope RMBench fosters principled progress toward scalable, memory-aware robotic manipulation.

%% file: sec/7-appendix.tex
\newpage
\appendices
\section{RMBench Task Descriptions}
\label{app:benchmark_description}

\begin{table}[h]
\centering
\small
\caption{\textbf{Task descriptions of the RMBench benchmark.}}
\begin{tabular}{@{}>{\itshape\centering\arraybackslash}p{0.28\columnwidth} >{\raggedright\arraybackslash}p{0.62\columnwidth}@{}} 
\toprule
\textbf{Task} & \textbf{Description} \\ 
\midrule 

Observe and Pick Up &
The robot observes a reference object, remembers it after it is hidden, and picks up the matching object from the table. \\ 
\midrule

Rearrange Blocks &
The robot moves a middle block to a pad, presses a button, and moves another block back to the middle position. \\ 
\midrule

Put Back Block &
The robot moves a block to the center, presses a button, and returns the block to its original pad. \\ 
\midrule

Swap Blocks &
The robot swaps two blocks using an empty pad and then presses the confirmation button. \\ 
\midrule

Swap T &
The robot swaps the positions and orientations of two colored T-shaped blocks. \\ 
\midrule

Battery Try &
The robot repeatedly tries different battery insertion orders and orientations until both batteries are correctly inserted. \\ 
\midrule

Blocks Ranking Try &
The robot repeatedly arranges colored blocks and presses a button until the correct ordering is found. \\ 
\midrule

Cover Blocks &
The robot covers blocks from left to right, then uncovers them in the red--green--blue order. \\ 
\midrule

Press Button &
The robot presses two buttons according to the shown digits and then presses the confirmation button. \\

\bottomrule
\end{tabular}
\label{tab:benchmark_description}
\end{table}

\section{Stability and Transferability of Task Memory Complexity}
\label{app:tmc_stability}

This section examines two properties of the Task Memory Complexity (TMC) protocol: whether it applies to tasks outside RMBench, and whether annotators other than the authors reproduce the intended labels. Both are necessary if TMC is to serve as a benchmark design criterion rather than as a post-hoc description of the tasks we happened to build.

\paragraph{Transferability to real-robot tasks}
TMC is defined by the structure of the task, namely the minimum number of past observations that resolve its ambiguous decision points, and not by any policy architecture. It is therefore applicable wherever those decision points can be enumerated. To test this, we annotated 50 manipulation tasks drawn from the DROID and AgiBot World datasets. Annotation followed the protocol of Sec.~\ref{sec:tmc}: enumerate the decision points at which the current observation admits more than one admissible action, identify the past observations that resolve each ambiguity, and discard redundant items. Every task in this set received a label within the existing space $\{M(0), M(1), M(n)\}$, and none required extending the label space or introducing task-specific exceptions. We report this as evidence of applicability rather than as a survey: the sample was not drawn so as to be representative of either dataset, and we therefore do not report a distribution over TMC levels for it. The claim supported by this exercise is that the protocol terminates with a well-defined label on tasks it was not designed around.

\paragraph{Annotation stability study}
We further conducted a study with ten independent participants who had not previously read the paper. Each participant received only the written protocol of Sec.~\ref{sec:tmc} and the three-way label space $\{M(0), M(1), M(n)\}$, and then annotated the same task set without consulting the authors or one another. Performance is measured as the per-annotator fraction of tasks whose label matches the reference label assigned by the task designers. As shown in Table~\ref{tab:tmc_user_study}, agreement averaged 96\% over the ten participants, ranging from 100\% to 80\%. For calibration, a labeler guessing uniformly at random over the three-way space would reach 33\%.

\begin{table}[t]
\centering
\small
\caption{\textbf{Stability of TMC annotation.} Ten independent participants labeled the same task set using only the written protocol of Sec.~\ref{sec:tmc}. Agreement is the fraction of tasks matching the reference labels of the task designers; Average, Highest, and Lowest are taken over the ten participants.}
\begin{tabular}{@{}lccc@{}}
\toprule
Metric & Average & Highest & Lowest \\
\midrule
Agreement (\%) & 96 & 100 & 80 \\
\bottomrule
\end{tabular}
\label{tab:tmc_user_study}
\end{table}

\paragraph{Discussion}
Two conclusions follow, both deliberately narrow. First, TMC applies to tasks it was not designed around, provided their decision points are enumerable; it does not provide a procedure for tasks whose success criterion is itself ambiguous. Second, participants working from the written protocol alone recovered the intended labels in the large majority of cases, which indicates that the protocol is transmissible in writing and does not depend on tacit knowledge held by the authors. This is the property required for others to extend RMBench with new tasks.

We state three limitations explicitly, since they bound how the numbers above should be read. First, the reported quantity is agreement with a reference standard rather than chance-corrected agreement among annotators. Fleiss' $\kappa$ over the participants would be the stronger measure, since it separates genuine consensus from the agreement expected by chance and does not privilege the authors' labels as ground truth; computing it requires retaining the full per-task label matrix, which we leave to an expanded study. The 96\% figure should therefore be read as a measure of how faithfully the protocol transfers to new annotators, and not as an accuracy or as evidence that the labels are independent of the authors' judgment. Second, the label space was closed and given to participants in advance, so the study measures consistency of application within that space rather than whether the space itself is the right abstraction. Third, the study reports aggregate agreement only, without a per-label breakdown. The $M(1)$ versus $M(n)$ boundary is the most delicate judgment the protocol requires, since it asks how finely a task decomposes into distinct semantic events, and a per-label breakdown would be needed to establish whether residual disagreement concentrates there.

\section{Visualization of Typical Baseline Errors}

\begin{figure*}[t]
    \centering
    \includegraphics[width=0.98\textwidth]{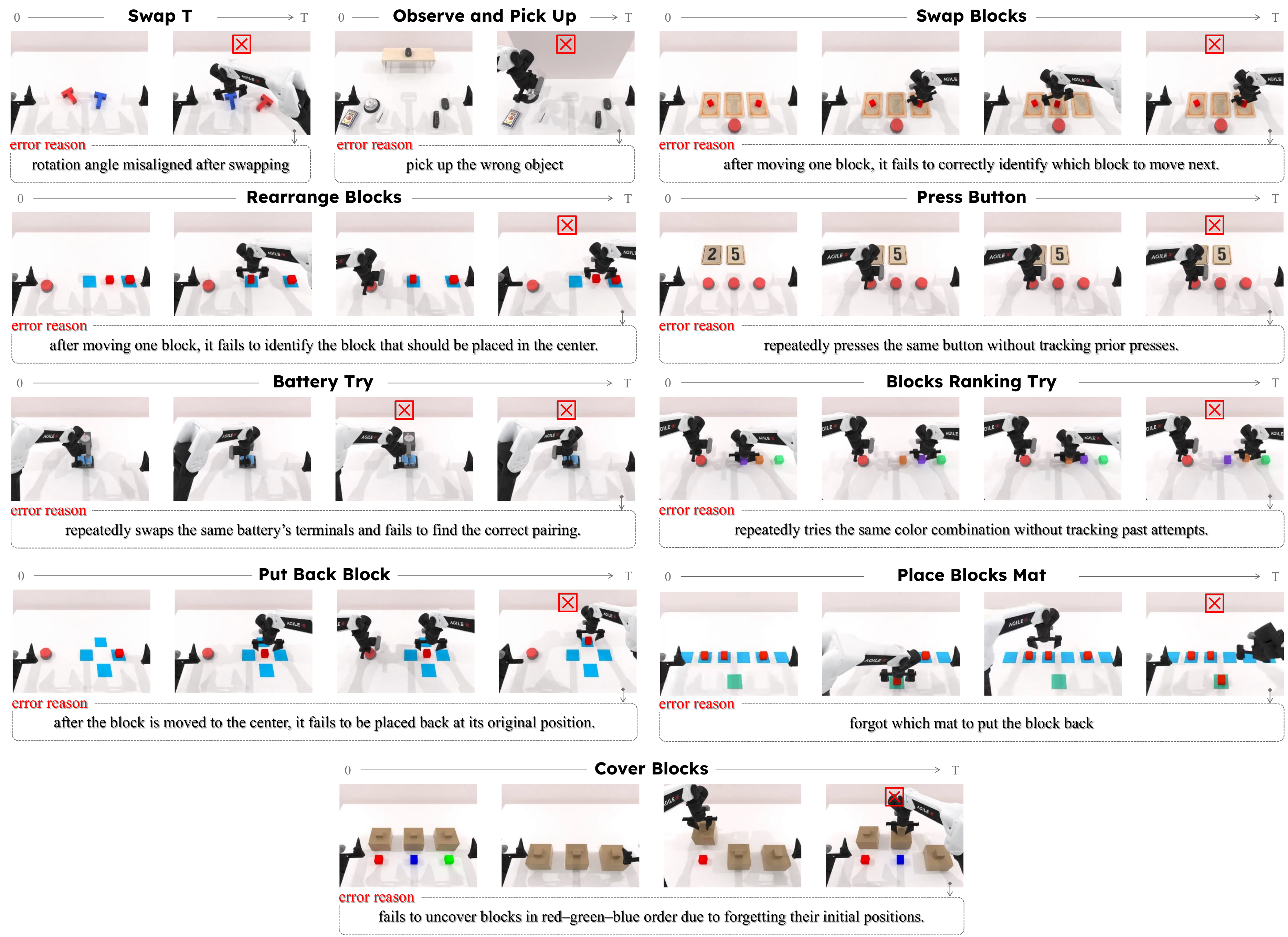}
    \caption{\textbf{Visualization of typical baseline errors.} Because these baselines predict the next action from the current observation alone, they cannot act reliably on non-Markovian tasks in which the required information is no longer visible.}
    \label{fig:baseline_error}
    \vspace{-2mm}
\end{figure*}

\section{Training Details}
\label{app:training_details}

\subsection{Planning Module}

In the Planning Module, we fine-tune the vision-language model (Qwen3-VL-8B-Instruct) using LoRA via LlamaFactory~\cite{zheng2024llamafactory} to enable reasoning over key memories. After fine-tuning, we deploy the model with vLLM~\cite{kwon2023efficient} for efficient loading and inference. The key hyperparameters used for VLM fine-tuning are summarized in Table~\ref{tab:planning_hyperparameter}. Training is conducted on 8 NVIDIA A800 GPUs and the duration of training for a single task is approximately half an hour.

\input{tables/planning_hyperparameters}

\subsection{Execution Module}
In this section, we detail the training infrastructure, training organization strategy, and hyperparameter configurations employed for the Execution Module in Mem-0.

\subsubsection{Model Configuration}

The Execution Module uses Qwen3-VL-2B-Instruct as its vision-language backbone $\mathcal{V}_{\text{exec}}$, which is smaller than the Qwen3-VL-8B-Instruct model used by the Planning Module, since $\mathcal{V}_{\text{exec}}$ is queried at every control step whereas $\mathcal{V}_{\text{plan}}$ is queried only at subtask boundaries. The action head is a DiT-B transformer trained from scratch with a flow-matching objective, using an action horizon of $H=30$ over a $16$-dimensional bimanual action space. The Subtask End Classifier operates on the same conditioning vector $\mathbf{c}_t$ as the action head; since $\mathbf{c}_t$ concatenates the two fused memory tokens with the text token, each of width $2048$, its dimension is $6144$, which the classifier maps through two residual blocks ($6144 \rightarrow 2048 \rightarrow 512$) and a linear head to a scalar logit. Table~\ref{tab:hyperparameters} reports the full configuration, including the memory hyperparameters (sliding window length $K$, anchor buffer size, and the two accumulation thresholds $L_{\text{sw}}$ and $L_{\text{mem}}$), the number of executed action steps $\Delta$, and the classifier loss terms ($w$, $\gamma_{\mathrm{f}}$, and $\lambda$). Two settings are worth highlighting. First, $\Delta = H = 30$, so each predicted chunk is consumed in full before the next inference call, which is what allows the module to sustain a 5--10~Hz control rate despite invoking a $2$B vision-language backbone. Second, the classifier is trained with a positive-class weight of $10.0$, since termination frames constitute a small minority of the training frames.

\subsubsection{Training Infrastructure and Time Budget}

The Execution Module of Mem-0 utilizes a single-task training strategy, where the model is trained from scratch for each specific task. Training is conducted on 8 NVIDIA A800 GPUs with a global batch size of 448 over 30K iterations. The duration of training for a single task is approximately 18 hours.

\subsubsection{Training Organization Strategy}

\textbf{Forward pass strategy.} Given the memory-centric architecture of Mem-0, we employ a specialized training methodology. Within each batch, the processes of VLM token generation and DiT-based action chunk generation are executed in parallel. Conversely, the fusion of the sliding and anchor memories requires the temporal integrity of the data, so this stage is processed serially, which keeps all frames within an episode sequentially aligned.

This arrangement keeps the VLM tokens usable for memory fusion without breaking episode ordering. We additionally maintain a global data structure during training that stores the memory state of each episode, which allows memory to be carried across batch boundaries.

\textbf{Dataloader implementation.} Consequently, the dataloader was custom-designed to align with this architecture. Episodes are distributed as evenly as possible across all GPUs. Each GPU processes its assigned episodes using a specific number of workers and manages the dataloader reset independently. Due to the stochastic nature of the distribution, as training iterations progress, the frames within a global batch become temporally desynchronized. This allows the model to learn simultaneously from data that span various time steps. 

\subsubsection{Hyperparameter Configurations}

Table~\ref{tab:hyperparameters} summarizes the key training hyperparameters. To balance the learning rate requirements of different modules, we implemented a grouped learning rate strategy alongside a cosine learning rate schedule with linear warm-up. Regarding numerical precision, the VLM and Memory Bank components operate in \texttt{bfloat16}, while all other modules utilize \texttt{float32}. Furthermore, regarding visual inputs, images are resized to $224 \times 224$ and subjected to mild data augmentation via frame-independent \texttt{ColorJitter}, aimed at enhancing the model's generalization capabilities.

\input{tables/execution_hyperparameters}

\section{Additional Visualizations and Analysis of Failure Cases in Mem-0}
\label{app:analysis}

Although Mem-0 improves substantially over the baselines, it remains far from solving RMBench. In this section we present representative cases in which Mem-0 fails, and identify the mechanism responsible for each, so that the residual error modes can be targeted directly by future work.

\subsection{Failure Analysis for $M(1)$ Tasks}

\begin{figure*}[tp]
    \centering
    \begin{minipage}[t]{0.4341\textwidth}
        \centering
        \includegraphics[width=\textwidth]{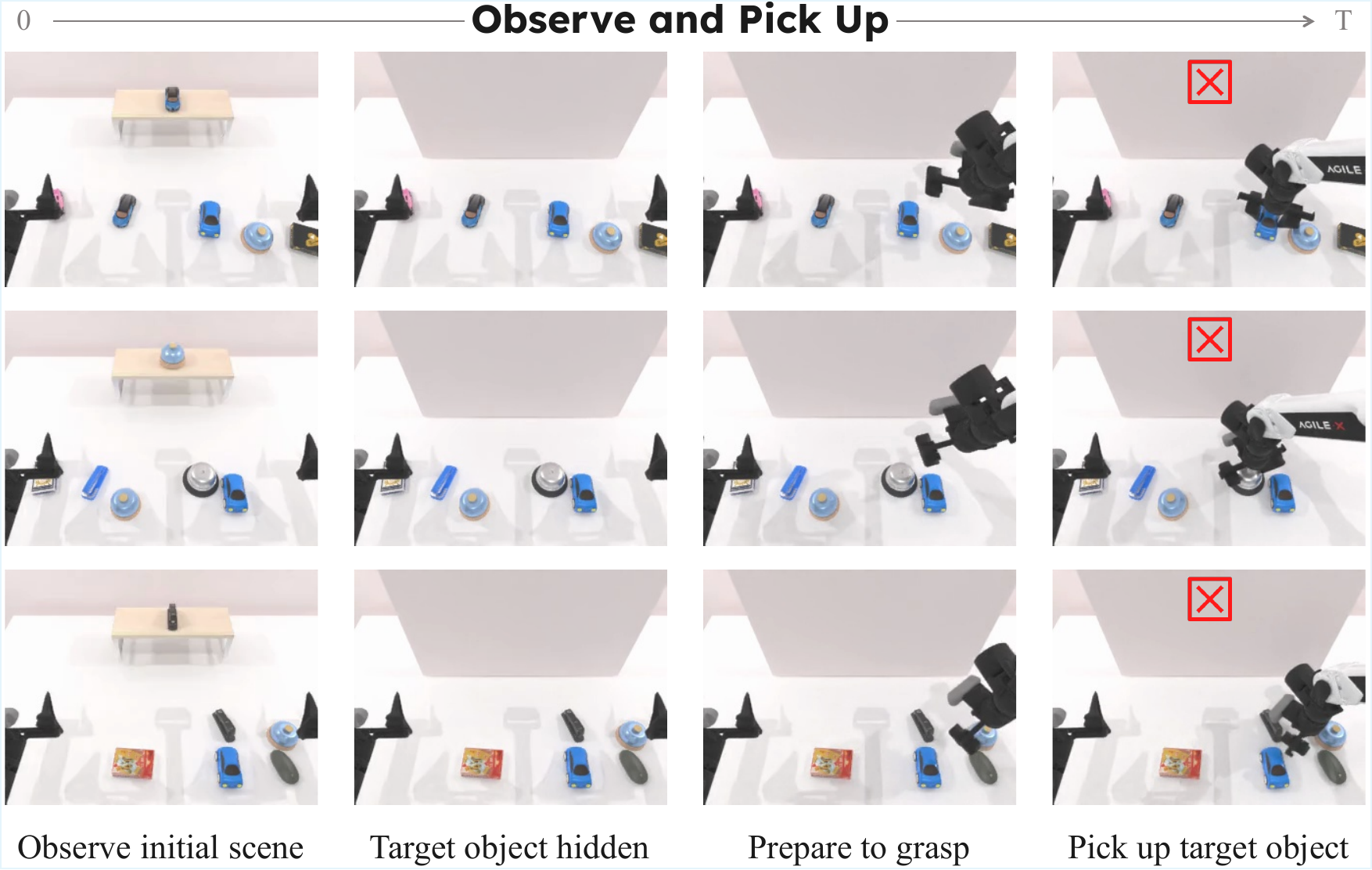}
        \caption{\textbf{Failure examples of Observe and Pick Up.} \textbf{(Top)} Confused by objects with similar colors and shapes. \textbf{(Middle)} Confused by identical object morphologies. \textbf{(Bottom)} General failure to identify the target, causing the gripper to reach a position averaged over the candidate objects rather than any one of them.}
        \label{fig:failure_observe_and_pickup}
    \end{minipage}
    \hfill
    \begin{minipage}[t]{0.5462\textwidth}
        \centering
        \includegraphics[width=\textwidth]{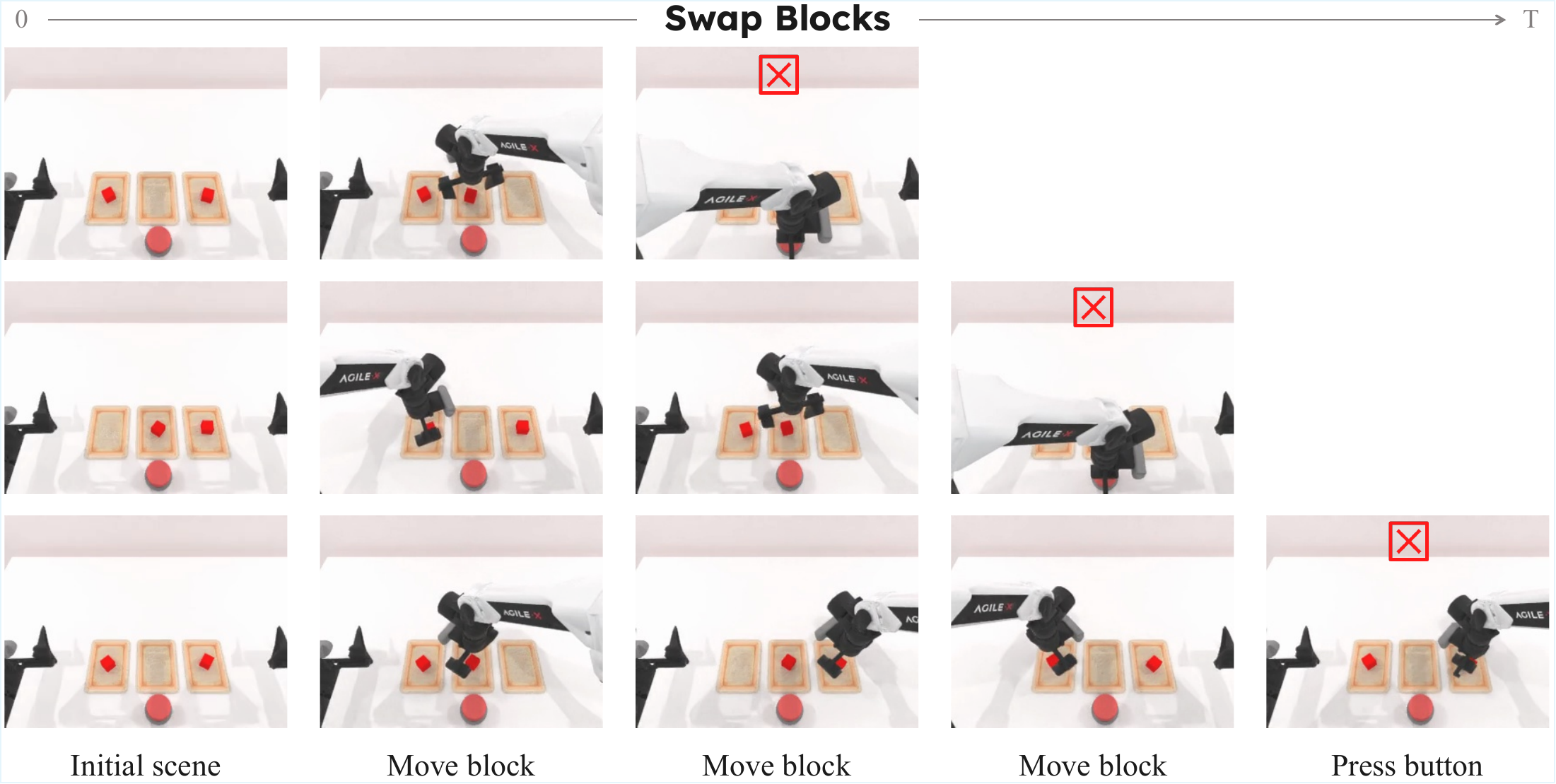}
        \caption{\textbf{Failure examples of Swap Blocks.} \textbf{(Top)} Premature termination after a single subtask. \textbf{(Middle)} Premature termination after two subtasks. \textbf{(Bottom)} Failure to terminate on time, resulting in the initiation of a redundant subtask.}
        \label{fig:failure_swap_blocks}
    \end{minipage}
\end{figure*}

\begin{figure}[htbp]
\centering
\includegraphics[width=\columnwidth]{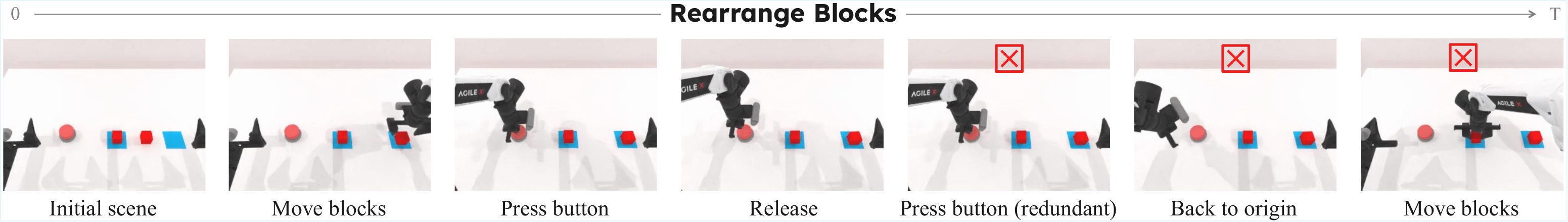}
    \vspace{-0.5em}
    \caption{\textbf{Failure examples of Rearrange Blocks.} Mem-0 redundantly presses the button, resulting in task failure.}
    \label{fig:failure_rearrange_blocks}
\end{figure}

For $M(1)$ tasks, in addition to the failures illustrated in Fig.~\ref{fig:baseline_error}, we summarize the representative errors encountered by Mem-0 below.

\textbf{Observe and Pick Up and Swap Blocks.} Fig.~\ref{fig:failure_observe_and_pickup} illustrates typical failure modes in the \textit{Observe and Pick Up} task, where Mem-0 fails to accurately identify the target object. Similarly, Fig.~\ref{fig:failure_swap_blocks} presents examples of misjudgments regarding the termination of the swapping sequence in the \textit{Swap Blocks} task, resulting in the confirmation button being pressed at inappropriate times. 

These instances indicate that the anchor memory influences the policy throughout the entire subtask horizon rather than only at its beginning, so the model must modulate how strongly that fixed context contributes to action prediction as the subtask proceeds. This is consistent with the observation in Sec.~\ref{sec:exp_on_ablation} that a persistent context can interfere with closed-loop visual alignment once the relevant history is already available elsewhere.

Nevertheless, the ablations confirm that anchor memory is the decisive component on $M(1)$ tasks, where its removal lowers the average success rate from 52.8\% to 26.8\%. The effect is most pronounced on \textit{Put Back Block} (90\% to 35\%) and \textit{Swap Blocks} (67\% to 15\%), the two tasks in which the episode-initial configuration is the only record of the required target.

\textbf{Rearrange Blocks.} Fig.~\ref{fig:failure_rearrange_blocks} illustrates failure cases in the \textit{Rearrange Blocks} task where Mem-0 performs excessive button presses, a behavior we attribute to the limitations of the sliding memory window. This is consistent with the ablations, in which removing the sliding memory reduces success on this task from 89\% to 62\%. Inspecting the rollouts, the frequency of redundant button presses increases markedly without sliding memory, which identifies it as a primary failure mode. The sliding memory therefore contributes substantially on this task while leaving room for refinement.

\textbf{Summary.} Two directions follow. First, richer fusion between the anchor and sliding memories would let the policy weight the persistent context according to how far the subtask has progressed, which is precisely what the failures above require and what a single-slot anchor read cannot express. Second, the residual errors on \textit{Observe and Pick Up} are perceptual rather than mnemonic, so they call for stronger visual grounding in the backbone rather than additional memory.

\subsection{Failure Analysis for $M(n)$ Tasks}

On $M(n)$ tasks the dominant failure mechanism is the accuracy and robustness of the Subtask End Classifier, consistent with the oracle comparison of Sec.~\ref{sec:exp_on_ablation}, in which ground-truth termination signals raise the $M(n)$ average from 28.5\% to 45.3\%.

\begin{figure}[htbp]
\centering
% \vspace{-6mm}
\includegraphics[width=\columnwidth]{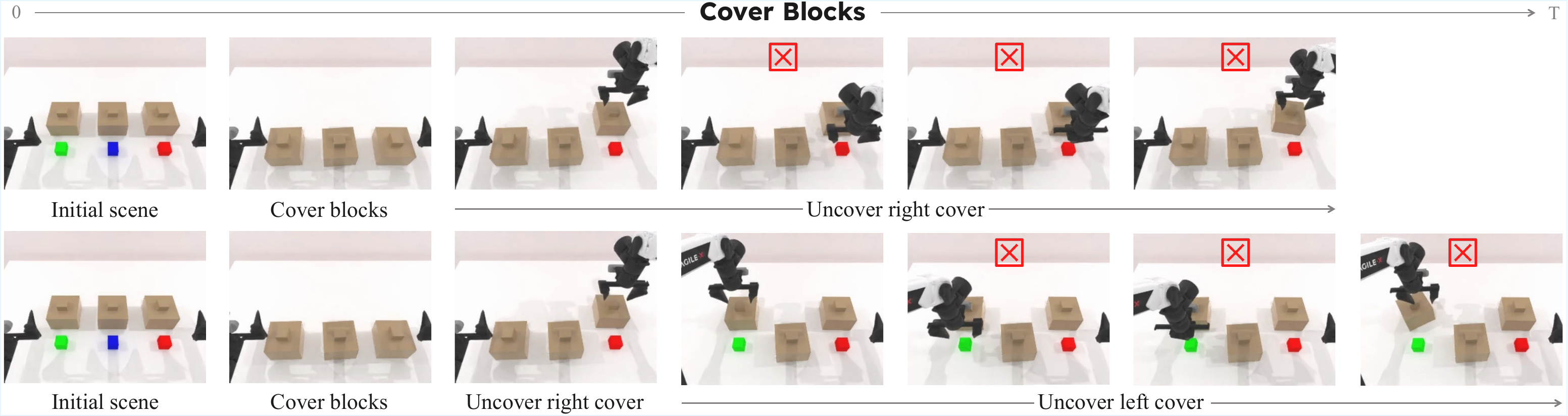}
    \caption{\textbf{Failure examples of Cover Blocks.} The classifier fails to detect the completion of an \textit{Uncover $\langle$color$\rangle$} subtask, so no subtask transition is triggered. Because the instruction remains unchanged, the policy continues to operate under a stale task context, leading to unintended and erratic behavior.}
    \label{fig:failure_cover_blocks}
\end{figure}

\textbf{Cover Blocks.} The classifier occasionally fails to resolve the current task progress, and in particular cannot discriminate whether a given state marks the beginning or the end of a subtask, since the two are visually similar for this task. Fig.~\ref{fig:failure_cover_blocks} shows an instance during \textit{Cover Blocks}.

\begin{figure}[htbp]
\centering
% \vspace{-6mm}
\includegraphics[width=\columnwidth]{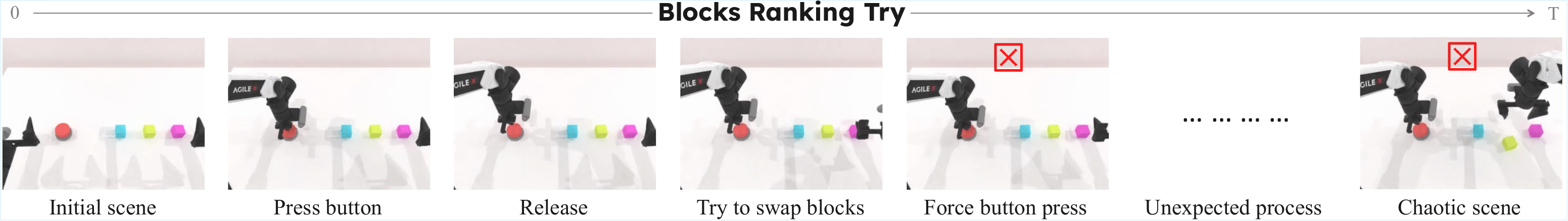}
    \caption{\textbf{Failure examples of Blocks Ranking Try.} Upon pressing the button, the system is expected to transition to the next subtask to execute the swapping of designated blocks. However, the classifier fails to trigger this transition promptly, causing the task to stall in the \texttt{Press button} state. This leads to a coordination conflict between the dual arms: the right hand attempts to initiate manipulation while the left hand remains tethered to the button-pressing instruction.}
    \label{fig:failure_block_ranking_try}
\end{figure}

\textbf{Blocks Ranking Try.} A second challenge concerns button-pressing operations. We observe that introducing button presses interferes with composite tasks that are not exclusively about pressing. In \textit{Blocks Ranking Try}, for instance, the transition between button-pressing and block-swapping is occasionally not carried out smoothly, as illustrated in Fig.~\ref{fig:failure_block_ranking_try}.

The task is also unforgiving: its success criterion requires the full ranking sequence to be correct, so a single execution error causes the episode to fail. This is consistent with the low absolute scores every method obtains on this task, and with the fact that supplying oracle termination signals raises Mem-0 from 18\% to 45\%, the largest such improvement among the $M(n)$ tasks.

\begin{figure}[htbp]
\centering
% \vspace{-6mm}
\includegraphics[width=0.85\columnwidth]{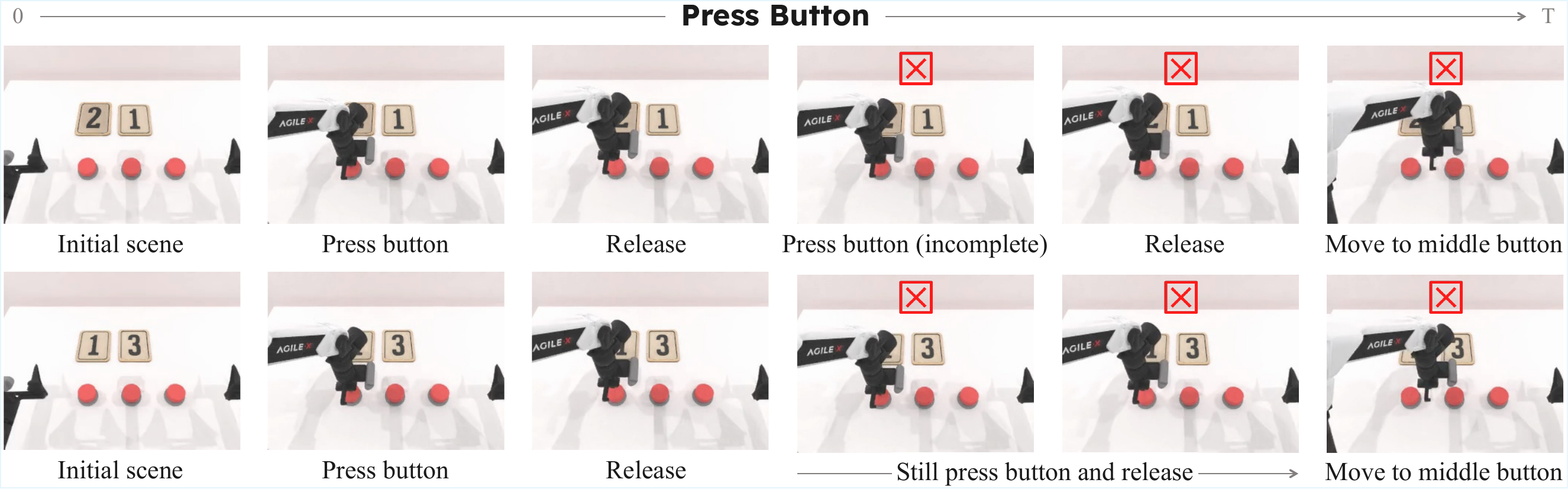}
    \caption{\textbf{Failure examples of Press Button.} \textbf{(Top)} Insufficient presses: the classifier issues a false positive termination signal even though the press was unsuccessful. \textbf{(Bottom)} Excessive presses: the classifier fails to recognize a successful completion, leading to redundant execution of the same subtask.}
    \label{fig:failure_press_button}
\end{figure}

\begin{figure}[htbp]
\centering
% \vspace{-6mm}
\includegraphics[width=\columnwidth]{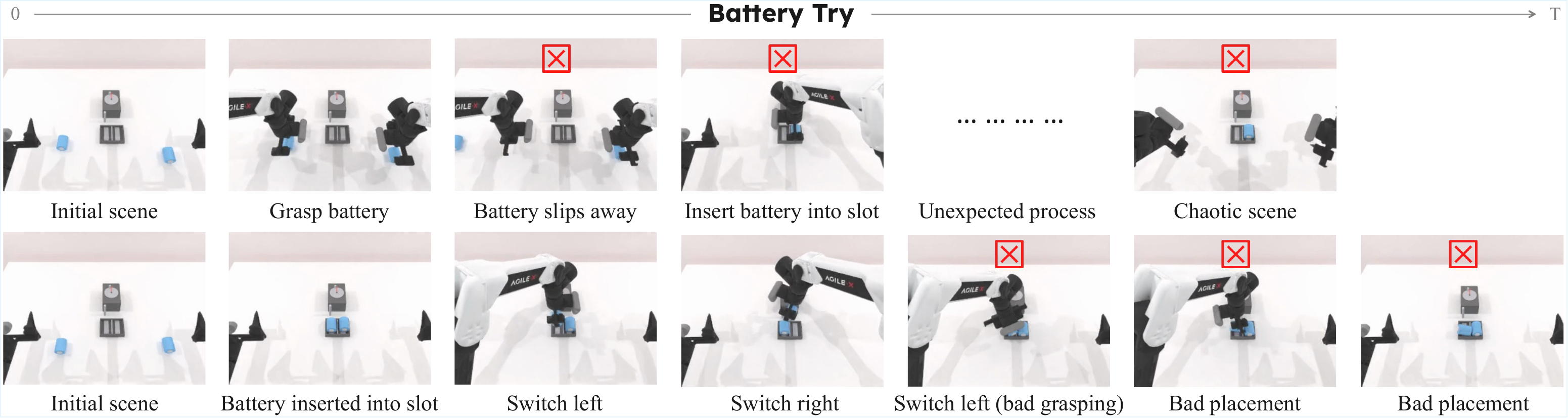}
    \caption{\textbf{Failure examples of Battery Try.} \textbf{(Top)} For the horizontally oriented cylindrical battery, a suboptimal grasp pose prevents a successful lift and causes significant displacement, leading the model into unforeseen observational states. \textbf{(Bottom)} The model fails to commit to a single manipulation strategy during battery adjustment, producing an action averaged over the competing strategies and hence an improper placement in the slot.}
    \label{fig:failure_battery_try}
\end{figure}

\textbf{Press Button.} The \textit{Press Button} task isolates this operation and therefore exposes its difficulty most clearly, which is why every policy in Table~\ref{tab:benchmark} scores zero on it. The button state, pressed versus unpressed, is reflected in the visual input only through very subtle cues, so the visual tokens produced by the VLM backbone do not resolve it. This places a bottleneck upstream of the classifier that no amount of memory can compensate for. Fig.~\ref{fig:failure_press_button} shows representative cases.

\textbf{Battery Try.} For \textit{Battery Try} the representative failure mode is limited manipulation precision rather than memory. It comprises insertion errors when placing the battery into the slot, and difficulty selecting a grasp strategy because the slot itself presents very subtle visual cues, as illustrated in Fig.~\ref{fig:failure_battery_try}.

\textbf{Summary.} Two directions follow from these observations. First, since several of the failures above trace to visual cues that are too subtle to survive tokenization, proprioceptive or tactile feedback would supply the classifier with information that vision does not carry. Second, because the classifier sits at the end of the pipeline and consumes $\mathbf{c}_t$ directly, improving upstream token extraction and the fusion of the anchor and sliding memories would improve termination detection without altering the classifier itself. Both directions target the bottleneck identified in Sec.~\ref{sec:exp_on_ablation} rather than the memory representation.

The current design nonetheless yields clear gains on $M(n)$ tasks, most visibly on \textit{Cover Blocks}, where Mem-0 reaches 68\% against 40\% for the strongest baseline.

%% file: tables/planning_hyperparameters.tex
\begin{table}[htbp]
    \centering
    \small
    \caption{\textbf{Hyperparameters for fine-tuning the Mem-0 Planning Module.}}
    \label{tab:planning_hyperparameter}
    \begin{tabular}{@{}p{0.50\columnwidth} p{0.42\columnwidth}@{}}
        \toprule
        \textbf{Configuration} & \textbf{Value} \\
        \midrule
        VLM Backbone $\mathcal{V}_{\text{plan}}$ & Qwen3-VL-8B-Instruct \\
        Fine-tuning Type & LoRA \\
        LoRA Rank & 8 \\
        Batch Size & 16 \\
        Learning Rate & $1.0 \times 10^{-4}$ \\
        Epochs & 25 \\
        LR Scheduler & Cosine \\
        Warmup Ratio & 0.1 \\
        Dtype & bf16 \\
        \bottomrule
    \end{tabular}
\end{table}

%% file: tables/execution_hyperparameters.tex
\begin{table*}[t]
    \centering
    \small
    \caption{\textbf{Hyperparameters for training and deploying the Mem-0 Execution Module.}}
    \label{tab:hyperparameters}
    \begin{minipage}[t]{0.47\textwidth}
    \vspace{0pt}
    \centering
    \begin{tabular}{@{}>{\raggedright\arraybackslash}p{0.56\columnwidth} >{\raggedright\arraybackslash}p{0.39\columnwidth}@{}}
        \toprule
        \textbf{Configuration} & \textbf{Value} \\
        \midrule

        \multicolumn{2}{@{}l}{\textit{General Training Settings}} \\
        \midrule
        Batch Size (global / per GPU) & 448 / 56 \\
        Iterations & 30,000 \\
        Max Grad. Norm & 2.5 \\
        LR Scheduler & Cosine \\
        Warmup Ratio & 0.05 \\
        Optimizer & AdamW \\
        Momentum & $\beta_{1}, \beta_2 = 0.9, 0.999$ \\
        Weight Decay & 0.005 \\
        Workers per GPU & 2 \\
        Random Seed & 42 \\
        \midrule

        \multicolumn{2}{@{}l}{\textit{Learning Rates}} \\
        \midrule
        LR (Base) & $1.0 \times 10^{-5}$ \\
        LR (VLM) & $1.0 \times 10^{-5}$ \\
        LR (Action Head) & $1.0 \times 10^{-4}$ \\
        LR (Classifier) & $1.0 \times 10^{-4}$ \\
        Min LR (Base) & $1.0 \times 10^{-6}$ \\
        Min LR (VLM) & $1.0 \times 10^{-6}$ \\
        Min LR (Action Head) & $5.0 \times 10^{-6}$ \\
        Min LR (Classifier) & $5.0 \times 10^{-6}$ \\
        \midrule

        \multicolumn{2}{@{}l}{\textit{Visual Input Settings}} \\
        \midrule
        Image Resize & $224 \times 224$ \\
        Image Aug. & ColorJitter\textsuperscript{\dag} \\
        \bottomrule
        \multicolumn{2}{@{}p{0.92\columnwidth}@{}}
        {\scriptsize \textsuperscript{\dag} ColorJitter parameters: $(0.1, 0.1, 0.1, 0)$.}
    \end{tabular}
    \end{minipage}
    \hfill
    \begin{minipage}[t]{0.47\textwidth}
    \vspace{0pt}
    \centering
    \begin{tabular}{@{}>{\raggedright\arraybackslash}p{0.56\columnwidth} >{\raggedright\arraybackslash}p{0.39\columnwidth}@{}}
        \toprule
        \textbf{Configuration} & \textbf{Value} \\
        \midrule

        \multicolumn{2}{@{}l}{\textit{Backbone and Memory}} \\
        \midrule
        VLM Backbone $\mathcal{V}_{\text{exec}}$ & Qwen3-VL-2B-Instruct \\
        Sliding Window Length $K$ & 30 \\
        Anchor Buffer Size & 1 \\
        Memory Attention Heads & 8 \\
        Memory Dropout & 0.1 \\
        Transition Threshold $L_{\text{sw}}$ (deploy) & 8 \\
        Buffer Reset Threshold $L_{\text{mem}}$ (train) & 8 \\
        \midrule

        \multicolumn{2}{@{}l}{\textit{Action Head (Flow Matching)}} \\
        \midrule
        Architecture & DiT-B \\
        Hidden Size / Layers & 2048 / 16 \\
        Action Dim. $d_a$ / State Dim. & 16 / 16 \\
        Action Horizon $H$ & 30 \\
        Executed Steps $\Delta$ & 30 \\
        Flow Time Dist. & $\mathrm{Beta}(1.5, 1.0)$ \\
        Flow Time Shift $s$ & 0.999 \\
        Inference Euler Steps & 8 \\
        Control Frequency & 5--10~Hz \\
        \midrule

        \multicolumn{2}{@{}l}{\textit{Subtask End Classifier}} \\
        \midrule
        MLP Widths & $6144 \rightarrow 2048 \rightarrow 512$ \\
        Dropout & 0.1 \\
        Positive Weight $w$ & 10.0 \\
        Focal Exponent $\gamma_{\mathrm{f}}$ & 1.0 \\
        Decision Threshold & 0.5 \\
        Loss Weight $\lambda$ & 0.2 \\
        \bottomrule
    \end{tabular}
    \end{minipage}
\end{table*}